\def\year{2018}\relax
\documentclass[letterpaper]{article} 
\usepackage{aaai18}  
\usepackage{times}  
\usepackage{helvet}  
\usepackage{courier}  
\usepackage{url}  
\usepackage{graphicx}  
\begin{document}
%
\title{Exploring Gameplay With AI Agents}
\author{Fernando de Mesentier Silva\\
New York University\\
Game Innovation Lab\\
Brooklyn, NY 11201\\
\And
Igor Borovikov\\
Electronic Arts\\
209 Redwood Shores Pkwy\\
Redwood City, CA 94065\\
\And
John Kolen\\
Electronic Arts\\
209 Redwood Shores Pkwy\\
Redwood City, CA 94065\\
\AND
Navid Aghdaie\\
Electronic Arts\\
209 Redwood Shores Pkwy\\
Redwood City, CA 94065\\
\And
Kazi Zaman\\
Electronic Arts\\
209 Redwood Shores Pkwy\\
Redwood City, CA 94065\\
}
\maketitle
\begin{abstract}

The process of playtesting a game is subjective, expensive and incomplete. In this paper, we present a playtesting approach that explores the game space with automated agents and collects data to answer questions posed by the designers. Rather than have agents interacting with an actual game client, this approach recreates the bare bone mechanics of the game as a separate system. Our agent is able to play in minutes what would take testers days of organic gameplay. The analysis of thousands of game simulations exposed imbalances in game actions, identified inconsequential rewards and evaluated the effectiveness of optional strategic choices.  Our test case game, The Sims Mobile, was recently released and the findings shown here influenced design changes that resulted in improved player experience.
\end{abstract}

\section{Introduction}
Player engagement is a crucial part of any game. Immersing the players in the game experience not only results in longer playing sessions, it also keeps them interested in coming back to the game. In contrast, experiencing inconsistent behaviors or an unnatural cycle of actions can result in early churn from the game. 

To ensure the game provides players with the intended experiences, designers conduct playtesting sessions. Playtesting consists of having a group of players experimenting with the game during development. Through it, designers not only want to measure the engagement of players, they  also want to understand how interactions reflect in their system. One main point of playtesting is to discover elements and states that result in undesirable outcomes. As a game goes through the various stages of development, it is essential to continuously iterate and improve. Relying exclusively on playtesting conducted by humans can be costly. Employing automated agents could reduce development costs through faster play sessions and the thorough exploration of the game space in much shorter time. These benefits can become even more valuable as a game grows in size, increasing the space of possible actions. Automated agents are also capable of playing the game trying to mimic the same decisions multiple times in order to generate statistically significant results.

Unexpected scenarios are not the only focus of playtests. During iterating on the game mechanics and looking for an ideal tunning, playtesting is used to gauge the impact of changes made to the game. Determining if changes had the desired impact on the gameplay or comparing progression over different builds of the game can also be made easier with the use of AI Agents. Agents can explore different routes through the game and statistical results can be used to compare the evolution of gameplay over different iterations.

In this paper, we present our work on using AI agents to facilitate the process of playtesting a game.  In the next section, we discuss related work. We then present The Sims Mobile, the focus of our game testing efforts. After this introduction, we elaborate our approach and the justifications for choosing it over a simpler, more straight forward, strategy of implementing the agent on the game client. We then show four use cases that were created following questions from the designers of the game. We later compare the success of our chosen approach in contrast with the game client approach. Lastly, we discuss our results, present our conclusions, and discuss future work.

\section{Related Work}


The concept of using agents to help playtest games has been previously explored by several researchers. De Mesentier Silva et al. investigated how using agents to automate playtesting could help designers of contemporary board games~\cite{de2017modernboardgameworkshopai,de2017contemporaryboardgameai,de2018EvolvingMaps}. With a focus on the game Ticket to Ride, their work presents four heuristic-driven agents tailor made to play the game. They present analyses that originate from the data gather by simulating the game with their agents and could help designers fine tune the game. The cases shown are interesting and informative, but it is hard to gauge with those were scenarios raised during development. Our approach was built to answer designer questions.

While we propose to use agents as a tool to help designers, other approaches to algorithmically address balance exist in the literature. Hom et al. presented AI techniques to optimize balance in abstract board games~\cite{[1]hom2007automatic}. Using a genetic algorithm, rules were searched in order to optimize for balance, represented by the number of draws and the advantage of going of going first. Krucher in turn looked to balance a collectible card game~\cite{[4]r2014}. By rating and modifying cards, AI agents would decide actions to take. Cards would be automatically modified following gameplay. Jaffe et al. analyze the contribution of several features in the game~\cite{[3]jaffe2012evaluating}. These features are automatically measured and tracked on a educational card game. Dormans evaluates the economy of a game and its impact on strategy~\cite{dormans2011simulating}. With the use of a framework designed to track the flow of resources enabling simulation and balance of games before they are built. Mahlmann et al. focuses on the card game Dominion, discussing how to achieve a balanced card set~\cite{[2]mahlmann2012evolving}. By playing the game with three different agents, intersections were found in the winning sets of different agents. These cards were deemed to be part of a set that contributed to a more balanced game, regardless of strategy. This work successfully demonstrates agent-based game balance evaluation.

Another dimension of research focuses on investigating approaches where AI and Machine Learning can play the role of a co-designer, making suggestions during development. This effort is called mixed initiative design~\cite{yannakakis2014mixed}. Liapis et al. presented a tool for creating real time strategy game maps~\cite{liapis2013sentient}. The tool, called Sentient Sketchbook, would make suggestions on how to change the proposed map design in order to optimize for a number of different features. Smith et al. presents a 2D platformer level designing framework~\cite{smith2010tanagra}. Users can create and manipulate key elements in the level and an algorithm proceeds to complete the level, guaranteeing its playability. Shaker et al. discusses Ropossum, a level design tool for the Cut the Rope game~\cite{shaker2013ropossum}. The work can address level playability, finish a level with incomplete design or generate a new one.

Other approaches have also touched on the contributions AI can make for the game design process. Browne et al. generates entirely new abstract games by means of evolutionary algorithms~\cite{browne2010evolutionary}. Games generated are measured in terms of predefined features of quality~\cite{browne2008automatic}. The most interesting designs generated were later then published. Salge et al. relates a games' design to the concept of relevant information with the use of an adaptive AI~\cite{[7]salge2010relevant}. Smith et al. addresses the behavior emanating from a design by having an engine capable of recording play traces~\cite{smith2010ludocore}. Nelson discusses alternate strategies to gather information from games, other than empirical playtesting~\cite{Metrics:IDP11}. Nielsen et al. relates quality in a game to the performance of multiple general game playing algorithms~\cite{nielsen2015general}. Isaksen automates playtesting, but with the goal of exploring the space of possible games represented from the concept of the game Flappy Bird. Variants with different game feel and difficulty are deemed interested and further explored~\cite{isaksen2015discovering,isaksen2015exploring}. De Mesentier Silva et al. searches the space of possible strategies for simple and effective introductory heuristics for playing blackjack and HULHE Poker~\cite{degenerating,de2018preflop,de2018post-flop}.


\section{The Sims Mobile}


{\em The Sims Mobile} is a mobile entry from the very popular game franchise The Sims. Gameplay focuses on ``emulating life'': players create avatars, called Sims, and conduct them through everyday activities. Common Sims actions can take range from cooking meals to going on dates with other avatars. The Sims Mobile is a ``sandbox'' game, where there is not a predetermined goal to achieve and instead players craft their own experience.


An important game mechanic revolves around managing your resource pool. Resources are required to perform actions. Each Sim has their own pool of resources. In addition to resources, actions also require time to be performed. After selecting an action for their avatar, such Sim is locked for the duration of it. Actions have a cool down restriction, meaning the same Sim can only repeat this action after such cool down is over. Lastly, resources are both recovered over time, as well as through the execution of specific actions.


Another core gameplay elements are events. To improve their relationships Sims have to complete specific in-game events. During an event, players have a set amount of time to complete specific actions in order to succeed. Each action rewards event experience (event XP) and the total XP acquired by the end measures the success of the Sim.


We conducted the reported work on development builds before public release.  The game received numerous updates since then, until it reached the current published product.

\section{Approach}

In order to playtest using AI agents, we need a model of the game mechanics. We originally attempted to implement AI agents controlling the actual game client. A series of limitations impeded this approach. First, the game mechanics could only be driven as fast as the client allowed. Since human gameplay was the only use case, it was built to fast enough to respond to human finger tapping, but no faster.  Second, neither graphics nor animation could be turned off, skipped, or otherwise bypassed. Rendering animations and user interface elements for our agent wasted computation time. Finally, the in-games menus could not be turned off as well, so the agents had to navigate them as part of their logic. It was also impossible to fast forward the time spent waiting for actions to be performed. These reasons, combined with having to rewrite the agent every time a new build of the game is released, forced us to search for a new approach.

The Sims Mobile uses a collection of JSON files to store game parameters subject to tuning. 
With the tunning files, we decided to re-implement the game mechanics as an application separate from the game client. This approach brought several advantages: complete control over the game state, graphics and UI avoidance, and the ability to advance the game clock when necessary. With the simulation of game mechanics, due to all the advantages listed above, gameplay runs at a much faster speed. The increase in speed is roughly thousand fold for  executing in-game actions. Despite having to re-implement the mechanics, we only need a slice of the game. Starting with the core system, we added more mechanics as needed for the analysis in question.

Fast simulation allowed for the application of search-based AI techniques. The system runs about a thousand actions every second, making it suited for future lookahead techniques such as A*. Given that all the agents in this paper are solving shortest path problems, A* would provide the globally optimal solution (and we exclude sub-optimal human-like behavior). This approach proved to be a more powerful technique than what we could explore by driving the game client. In the results section, we look further into the comparison between the two approaches.

\subsection{Using the A* Algorithm}

Our experiments were agreed upon in consultation with the game designers with whom we identified clear goals for the agents. Requirements to reach these goals are explicit and the rewards for each action usually have some direct impact on reaching the conditions. For this reason, we decided to use the A* algorithm to play the game. The challenge is then to build a heuristic that can target the gameplay style that we are looking for in an experiment.

Although the elements that influence the goal are clear, building a heuristic from them is not. Weighting the different components to achieve the desired outcome is not simple. The parts have to be managed delicately and minor changes can result in different strategies.

The experiment guides the heuristic function construction. We selected the experiments after meeting with the game testing team. They propose questions about the game tuning, such as imbalance or possible exploits. We then write a heuristic aimed at exploring the issues raised. This means heuristics are frequently changed or re-written. These heuristics target different in-game activities, and for each case, various elements are used to create effective functions.

Our experiments can require thousands of actions to be completed. To reduce the amount of time the agent spends to pick an action, we limit the amount of nodes A* can search. For our experiments, we limit the search to 2000 nodes for every action. Consequently A* might not find the optimal action to take. However, from testing we concluded that searching 2000 nodes achieved the desired results in all experiments, and had fast computational speed, with each decision taking at most one second.

Some actions in the game have extremely similar rewards. Difference in actions might be only in the cool down time. When applying the heuristic, we could have multiple game states with the same evaluation.
When nodes have the same priority, we randomly choose which one to evaluate.

\section{Use Cases}

In this session, we detail four different use cases that were assessed by having an A* agent playtest the game. First, we state the question raised by the designers. Then we proceed to give a detailed description of the in-game representation of the use case. Finally, we present the results and discuss how they influenced design decisions.

\subsection{Use Case: Relationship imbalance}

\begin{figure*}[th]
  \centering
  \includegraphics[width=1.0\linewidth]{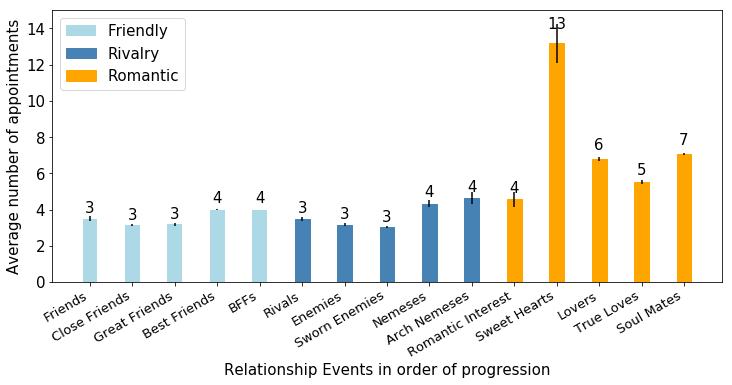}
  \caption{Bar chart showing the average number of actions (appointments) needed to complete each event. Events are color coded to reflect the relationship category they belong to. The number on the top of the bar shows the rounded value of each average. The black lines on top of the bars represent the variance. The Sweet Hearts (the second event of the Romantic category) appears as an outlier, requiring twice as many actions as any other event. This turned out not to be the tuning the game designers desired and was changed over the next iterations of the game.}
  \label{Figure:bar_char_relationships}
\end{figure*}

\textbf{Question:} {\em Is there a significant imbalance when comparing different relationship categories?}

For two Sims to develop a relationship, they need to pursue one of three mutually exclusive categories: Friendship, Romance, and Rivalry. Each relationship type requires completion of five events in a specific order, e.g., Friendship progresses through Friends, Close Friends, Great Friends, Best Friends, and finally, BFFs. An imbalance between categories implies that players need to put significantly more effort to pursue one category compared to another. The metric for this difference is the number of the actions required to complete all the events for the relationship category.

Our experiment measures how many actions are needed to complete the events in each category. Our objective is to compare the number of events between categories considering their order, e.g., compare the first events of category A with the first event of B and C. For this experiment, our heuristic rewards the agent for acquiring relationship experience points and for successfully completing relationship events. Our simulation stopped when the Sim completed the fifth, and last, event of any relationship category.

There are aspects that make exploring relationships unique compared to other experiments. The players do not select their relationship category by navigating a menu, like they do for careers and hobbies. Instead, the category of the first event chosen leads the relationship to that category. Another feature unique to relationships are actions with delayed effect dependent on the category. We do not include those in our heuristic for this experiment, causing the A* to reach local optimum depending on the randomization. To overcome such, we run the experiment 1000 times and take the average of those runs. Repeated sampling allowed the use of the same heuristic to explore all categories of relationship. Since the first event for all categories are very similar in requirements, whichever is first on the list of actions is selected. This randomization guarantees that we have a significant amount of samples for each category after 1000 runs.

Figure \ref{Figure:bar_char_relationships} shows the result of the experiment. While the agent achieved Friendship and Rivalry events with only three or four appointments, the second event of the Romantic category, required thirteen appointments. This outlier showcases that more effort was needed to progress in Romantic relationships. Following this find, designers analyzed the tuning data for the second event of each category and discovered the second event for the Romantic relationship to require about two times more experience than any other second event. The game team adjusted the number of appointments for this event over the next iterations of the game.

\subsection{Use Case: Time needed to progress in a career}

\begin{figure}[th]
  \centering
  \includegraphics[width=1.0\linewidth]{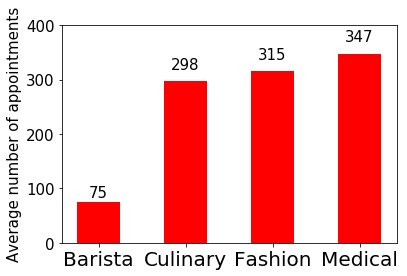}
  \caption{A bar chart comparing the number of actions (appointments) required to reach the goal in each career. Barista is much smaller because it can only go up to level 5, while the others are going up to level 10.}
  \label{Figure:4_career_bar_chart}
\end{figure}

\textbf{Question:} {\em How many actions are needed to progress in the careers?}

Players can pick one career for each of their Sims. Characters progress in their career by completing events specific to that profession. Each career has several levels to represent the progress made and the actions of the events reward career experience points that are used to reach higher levels. Each career has a maximum level to reach and may have different experience point requirements for their levels.

Answering this question would provide guidance to the game team on the relative balance of different careers. To perform this experiment, we examine the first four careers that are available to the players: Barista, Culinary, Fashion, and Medical. While the maximum level for Barista is 5, the others can reach level 20. For our experiment, we stopped when reaching the maximum level for Barista or level 10 on the others. We capped these careers as it provided sufficient data to establish the speed of progression in the early game.

The heuristic to progress in careers is similar to the relationship heuristic. Career level, career experience points and completing career events are direct rewards we look for. The Barista and Culinary careers also have a small extra mechanic: Sims need to perform an action to generate an object, such as coffee and tea for Barista and cooked dishes for Culinary, which then enable them to perform a second action that rewards experience points. For that reason we factor those objects into our heuristic as well.

Unlike relationships, players do not execute a character action to choose their career. They instead select from a menu, an action requiring none of the Sim's resources. We run separate experiments for each career, by assigning our desired career directly to the Sim at the start of the simulation. The same heuristic was used across all careers.

Figure \ref{Figure:4_career_bar_chart} shows the results of our experiment. Barista has less than a third of the actions required to complete the career, but it only goes up to level 5. Meanwhile, Culinary, Fashion and Medical all go up to level 10, but have a difference in number of necessary actions. The game team decided that these values represented their original design intentions and choose not to make any changes.

\subsubsection{Incorrect tuning discovered}

\begin{figure}[th]
  \centering
  \includegraphics[height=0.75\linewidth,width=1.0\linewidth]{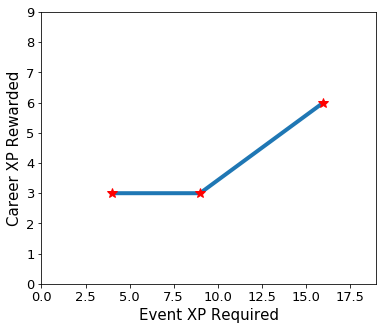}
  \caption{Line plot showing the relation between the amount of experience required to achieve each step of the event and the amount of career experience it rewards for reaching it. Each step in the event is marked by a red star. This plot evidences that reaching the second step takes more than double the effort of the first one for the exact same reward.}
  \label{Figure:career_event_star_incorrect_tuning}
\end{figure}

When running experiments for the Barista career, we noticed that agents made unexpected decisions in the case of one specific event. Usually the agent would progress the event until it was completed. For this specific event, however, it decided to do a handful of actions and then fast-forward the game time until the event timed out. Instead of receiving full reward for the event, it did only enough actions to receive the smallest reward the event could give.
When investigating the tuning of such event, we pinpointed the cause of this behavior. Figure \ref{Figure:career_event_star_incorrect_tuning} shows the cost-reward relation of this event. Reaching the second step of the event would have the agent do at least twice as many actions when compared to stopping at the first step, but it would still generate the same reward. When presented with this novel strategy, the game team traced it to an error in the tuning parameters and fixed it.

\subsection{Use Case: Effect of objects on careers}

\begin{table}[t]
\begin{center}
\begin{tabular}{ |c|c|c|c| } 
 \hline
	\textbf{Career} & \textbf{Actions Reduction (\%)} & \textbf{$\rho$ / Action Save} \\
 	\hline
 Barista & 5\% & 56.7 \\ 
 	\hline
 Culinary & 26\% & 45.8 \\ 
 	\hline
 Fashion & 20\% & 52.6 \\ 
 \hline
\end{tabular}
\end{center}
 \caption{Table showing how objects affect career progression. Here, $\rho$ denotes in-game resources available to all players. The ``Actions Reduction (\%)'' column shows the percentage of fewer actions required to achieve level threshold. The ``$\rho$ / Action Save'' column shows the ratio between the amount of resources used, for all objects affecting that career, and how many actions less they would need to take. The Medical career objects are not availabe below level 10.}
 \label{tbl1:object_career_table}
\end{table}

\textbf{Question:} {\em How does object acquisition impact career progress?}

When progressing through a career, players can acquire objects specific to that career. These objects unlock object-specific new event actions that have higher utility for faster progression than the regular actions. The objects become available as the Sim progresses through the career levels.

For this experiment, we wanted to gauge the impact of objects in the career progress. While we used the same heuristic for exploring careers, the simulation would now award objects to the agent the moment they were available without an exchange of resources. We then compared it to the number of actions needed to complete the same career goals we set out before, with no objects to use.

Table \ref{tbl1:object_career_table} shows the results of the experiment. Objects made a bigger impact on the Fashion and Culinary careers, while having little impact on the Barista. The Medical career shows no impact, since objects are only available past level 10. We also displayed the ratio of the amount of resources the players would have to exchange to acquire the objects by the amount of actions they would save. Designers analyzed the findings and  changed the tuning of the object's to increase their impact and make them more accessible.

\subsection{Use Case: Comparing the off-time between builds}

\begin{table}
\begin{center}
\begin{tabular}{ |c|c|c|c| } 
 \hline
	\textbf{Career} & \textbf{Evt Actions} & \textbf{Tot. Actions} & \textbf{Sessions} \\
 	\hline
 Barista A & 75 & 82 & 2 \\ 
 	\hline
 Barista B & 347 & 381 & 24 \\ 
 	\hline
 Culinary A & 298 & 327 & 8 \\ 
 	\hline
 Culinary B & 1506 & 1643 & 94 \\ 
 	\hline
\end{tabular}
\end{center}
 \caption{Table showing the comparison of the number of actions and sessions needed for Barista and Culinary careers on build A and build B. The Evt Actions column refers to how many event related actions were taken. The Tot. Actions column shows the total number of actions taken during the experiment. The Sessions column shows the number of sessions necessary to finish the experiment.}
 \label{tbl2:career_by_build_table}
\end{table}

\textbf{Question:} {\em Did gameplay change between builds?}

With major changes between game iterations, significant impact on the gameplay can be felt. The question ``how much impact did the changes have'' is on everyone's mind. Using our AI agent to play both builds of the game, we can create metrics for comparison. Since we are not simulating the full game experience, our approach can only make punctual statements, but those can point to the impact of changes.

For our experiment, we compare progression through the Barista and Culinary careers between two significantly different builds named A and B for simplicity. The resource management system is the key change between these builds, as a consequence we expect a big impact in the gameplay and the goals our agents want to achieve.


Table \ref{tbl2:career_by_build_table} shows the results of our experiments. The difference in the number of actions performed between builds A and B is evident. For both careers, build B requires over four times as many actions to reach the target goals. We also compared builds in terms of sessions. A session is a period that starts when players log into the game and ends when no more actions are available. We saw a significant increase in the number of sessions from the build A to B, but the length of the session also changed. While in build A, players had to wait in average six hours between sessions, in build B the average wait time drops to about 45 minutes. This way, playing build A allows faster progression, but build B incentivizes players to keep checking the game throughout the day to steadily improve their progress. The two different builds provide very different experiences. It is up to the designers to decide which experience they prefer.

\begin{figure}[t]
  \centering
  \includegraphics[width=1.0\linewidth]{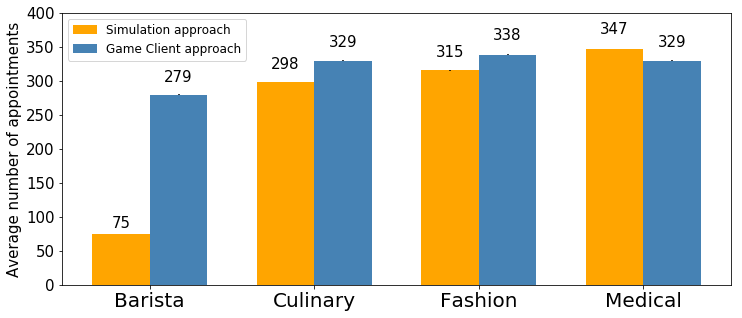}
  \caption{Bar chart comparing career progress in each approach. Numbers are the average amount of actions (appointments) to reach the goal. Our approach finds a path closer to optimal play in all but one career.}
  \label{Figure:career_approach_comp}
\end{figure}

\section{Comparing Approaches}

Earlier, we identified some of the limitations of running an agent on the game client. We now compare the results of the two different approaches, simulation based and  game client based, using the career experiment as the basis of our comparison. We are looking to compare how many actions each approach needs to achieve the career goal. The agents of each approach use different algorithms: while the simulation uses A*, the game client approach uses Softmax.\footnote{Namely, we used Softmax over utility of valid actions, trained with stochastic gradient ascent to optimize linear weights of the action parameters. 
During the model execution, lower temperature reduced the variance of the results.
}

Figure \ref{Figure:career_approach_comp} shows the comparison between the two approaches. The simulation based approach reaches a style of gameplay closer to optimal in all but one case. Optimal in this case would be using the minimum amount of actions. The Barista stands out for the large difference. For the medical career, the 2000 node A* cutoff could be moving the algorithm into a local optimal, while the game client Softmax was closer to optimal play.

\begin{figure}[t]
  \centering
  \includegraphics[width=0.943\linewidth]{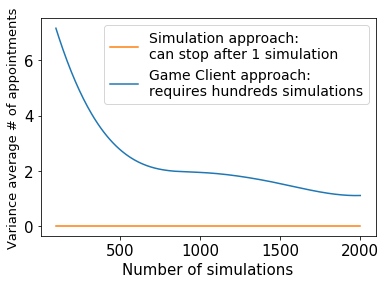}
  \caption{Comparing the variance between the two approaches. While A* on the simulation approach achieves deterministic play and has no variance, the Softmax game client approach has high variance and needs a considerable number of simulations to converge.}
  \label{Figure:approach_variance_comp}
\end{figure}

We can also compare the number of simulations needed for significant results. We ran 2000 simulations for each approach, and the conclusion is obvious. The simulation with A* agent achieves a deterministic playstyle, having no variance. In contrast, the game client Softmax agent has high variance requiring numerous simulations for convergence. Figure \ref{Figure:approach_variance_comp} compares the variance between the two approaches. A single run of A* already achieves our goals, which balances the time spent re-implementing the mechanics.

\section{Discussion and Conclusion}

We illustrated the advantages of AI-based playtesting in game development and how it can help designers to validate their work. We have shown the limitations of trying to implement an AI agent on the game client and proposed the approach of building a simulator of the game mechanics. Our approach gives us full control over the experiments and avoids the difficulties of coupling with an instrumented game client. We have also highlighted the power of this technique with four use cases proposed by the game team and which results were later presented to the designers, informing decisions they took to make changes.
%
%

Because most games keep their data in a standard format, it is becoming easier to write a simulation outside of the game client. These tools may come in existence even before a playable game client is built, even as the game is being designed. AI can then have an even bigger impact in playtesting, assisting from early stages in the game development, helping speed up the production cycle, saving time and efforts while achieving more balanced gameplay.
%
%

We have shown that our approach produces overall stronger results by empowering search-based algorithms. A basic algorithm such as A* achieves more precise results than agents running within the limitations of the game client.

A* delivered convincing results, but for the price of developing and tuning the heuristic function. The Monte Carlo Tree Search algorithm could be a viable alternative eliminating this overhead of making a new heuristic in favor of a custom win condition for each experiment. MCTS Agents were proven successful at gameplaying, and we believe it would be no different for a game such as The Sims Mobile.

\bibliography{bibliography}
\bibliographystyle{aaai}
\end{document}